\documentclass[verbose=true,letterpaper,margin=1in]{article}

\usepackage[final]{corl_2026} %

\usepackage{xcolor}
\usepackage{xspace} %
\usepackage{graphicx}
\usepackage{graphicx}
\usepackage{amsfonts} %
\usepackage{amsmath} %
\graphicspath{{figures/}}
\usepackage{float} %
\usepackage{algorithm}
\usepackage{algpseudocode}
\usepackage{booktabs}
\usepackage{bbm}
\usepackage{lipsum}
\usepackage{wrapfig}
\usepackage{enumitem} 
\usepackage{hyperref}
\hypersetup{
    colorlinks=true,
    linkcolor=blue,
    filecolor=magenta,      
    urlcolor=blue,
    pdftitle={Cloak: Zero-Shot Cross-Embodiment Manipulation by Masking the End-Effector from the VLA},
    pdfauthor={Michael Piseno, Guy Tevet, C. Karen Liu},
}

\usepackage{booktabs}
\usepackage{xcolor}
\usepackage{xspace}

\newcommand{\methodname}{\textit{Cloak}\xspace}
\newcommand{\vlaname}{Cloak-VLA\xspace}
\newcommand{\titlename}{Cloak: Zero-Shot Cross-Embodiment Manipulation by Masking the End-Effector from the VLA}

\newcommand{\baselinePiDroid}{\ensuremath{\pi_{0.5}}\textit{-droid}\xspace}
\newcommand{\baselineNoAug}{\vlaname{}-NoAug\xspace}
\newcommand{\baselineOverlap}{\vlaname{}-Overlap\xspace}
\newcommand{\baselineLAP}{LAP-VLA\xspace}
\newcommand{\baselinePi}{\ensuremath{\pi_{0.5}}\xspace}
\newcommand{\baselinePiDroidNoWrist}{\ensuremath{\pi_{0.5}}\textit{-droid}-NoWrist\xspace}

\newcommand{\val}[2]{#1\,{\scriptsize\textcolor{black!55}{$\pm$#2}}}
\newcommand{\ver}[1]{}
\usepackage{float} %
\usepackage{multirow} %

\newif\ifmodepaper \modepaperfalse
\newif\ifmodesupp  \modesuppfalse
\newif\ifmodefull  \modefulltrue

\ifmodesupp
\title{\titlename \\[0.4em] \textcolor{blue}{Supplementary Material}}
\else
\title{\titlename}
\fi

\author{
  Michael Piseno\textsuperscript{*} \quad Guy Tevet\textsuperscript{*} \quad C.~Karen Liu\\[0.3em]
  Stanford University\\[0.3em]
  \textsuperscript{*}Equal contribution\\[0.3em]
  \href{https://tml.stanford.edu/cloak/}{\texttt{tml.stanford.edu/cloak}}
}

\begin{document}
\maketitle

\ifmodesupp

\begin{figure}[H]
  \centering
  \includegraphics[width=\linewidth]{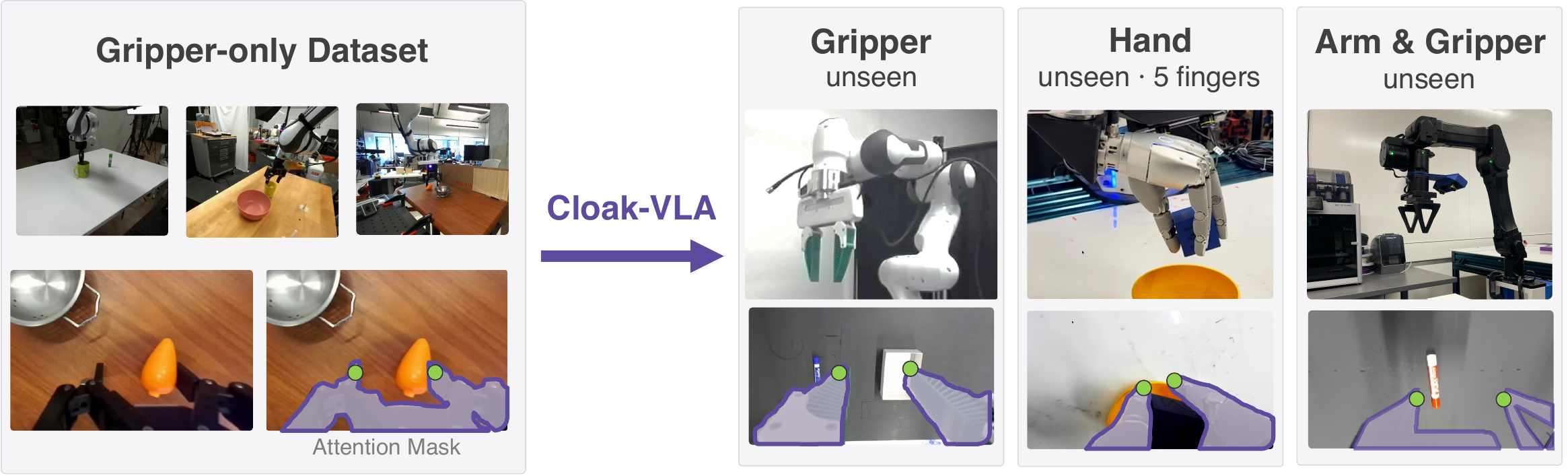}
  \vspace{-5pt}
\caption{
We cloak the end-effector from its own wrist camera, letting a VLA trained on a single parallel-jaw gripper transfer zero-shot to unseen embodiments, including a five-fingered hand.
}
\label{fig:teaser}
\end{figure}

\appendix

\begin{center}
  {\LARGE\bfseries Appendix\par}
\end{center}
\vspace{0.5em}

\section{Data Processing}
\label{sec:supp_data}

\subsection{Camera Extrinsics Estimation}
\label{subsec:supp_extrinsics}

The wrist-camera extrinsics shipped with DROID are noisy and not suitable for the pixel-level mask alignment needed by \methodname{}. We therefore re-estimate the 6-DoF camera pose in the end-effector frame, $T_{ec} \in SE(3)$, per episode, via a procedure we call Silhouette Calibration. Given $T_{ec}$ we can render masks in simulation. We note that $T_{ec}$ is optimized against our MuJoCo model, which may incur some error from small inconsistencies between our model and the actual robot setups in the DROID dataset. Moreover, MuJoCo's pinhole camera model does not exactly match the wrist camera's optics. In practice, we find these sources of error to be negligible.

\paragraph{Optimization.} To compute the wrist camera extrinsics, we solve the optimization problem

\begin{equation}
    T_{ec}^{*} = \arg\min_{T_{ec} \in SE(3)} \; 1 - \mathrm{IoU}\bigl(M(T_{ec}),\, M_{\text{gt}}\bigr),
\end{equation}

where $M_{\text{gt}}$ is a pseudo-ground-truth mask and $M(T_{ec})$ is the rendered mask from simulation using a candidate extrinsics pose. Because the renderer is non-differentiable, we use a zeroth-order optimizer (Nelder–Mead) over the 6-DoF parameterization of $T_{ec}$. The optimization is initialized from the mean 6-DoF camera pose in the original DROID dataset, $T_{ec}^{0}$.

\paragraph{Pseudo–ground-truth gripper mask.} We compute a pseudo-ground-truth mask to optimize $T_{ec}$. This pseudo-ground-truth mask is noisy and not suitable for \methodname{} directly, but provides reliable supervision for the optimization of $T_{ec}$. In particular, we exploit two properties of the DROID wrist view: (1) the gripper occupies roughly the same image region across episodes and (2) its pixels are consistently dark, with low temporal variance across frames compared to the non-gripper parts of each frame.

For each episode we take only gripper-open frames ($g < \tau_{\text{open}}$, assuming $g{=}0$ is open), convert the frames to grayscale, and crop the frames to a region in which the gripper always appears (the bottom-right). Within the region we compute two pixelwise scores across the frames: the temporal median intensity and the temporal standard deviation. Keeping the pixels below the median of each score yields two candidate masks, $M_{\text{int}}$ (dark) and $M_{\text{std}}$ (rigid). We take their intersection
\begin{equation}
    M_{\text{gt}} = M_{\text{int}} \cap M_{\text{std}},
\end{equation}
which rejects pixels that are merely dark (e.g. a dark background) or merely rigid (e.g. a solid-colored table in the same camera position the entire episode). The full Silhouette Calibration procedure is summarized in Algorithm~\ref{alg:supp_extrinsics}, and Figure~\ref{fig:supp_extrinsics} shows a representative frame before and after optimization.

\subsection{Additional Data Processing Details}
\label{subsec:supp_data_filtering}

\paragraph{Data filtering.} We filter certain episodes from the training dataset to improve data quality. First, following openpi~\cite{openpi}, we filter all episodes of the DROID dataset labeled as a failure, amounting to approximately 22\% of the 95K episodes in the dataset. We also filter episodes with extrinsics outliers from Silhouette Calibration. Specifically, we remove the episodes whose optimized camera extrinsics are the furthest 1\% from $T_{ec}^{0}$ in either translation or rotation (geodesic), accounting for approximately 1.7\% of the remaining 75K successful episodes. Finally, of the remaining episodes after these two filtering steps, 5\% is held out as a validation set. The final training set contains approximately 70K episodes.

\begin{figure}[t]
    \centering
    \includegraphics[width=\linewidth]{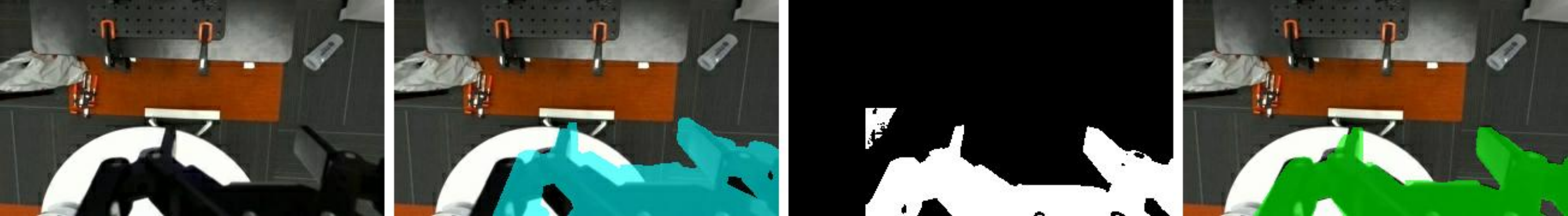}
    \vspace{-1.5em}
    \caption{Wrist camera extrinsics estimation on a representative DROID frame. From left to right: the raw wrist RGB frame, the simulated gripper mask rendered at the initial extrinsic $T_{ec}^{0}$ (cyan) overlaid on the RGB, the pseudo-ground-truth mask $M_{\text{gt}}$; and the simulated gripper mask rendered at the optimized extrinsic $T_{ec}^{*}$ (green) overlaid on the RGB. The optimized mask closely tracks the real gripper, despite a noisy pseudo-ground-truth.}
    \label{fig:supp_extrinsics}
\end{figure}

\begin{algorithm}[t]
\caption{Silhouette Calibration: per-episode wrist camera extrinsics refinement.}
\label{alg:supp_extrinsics}
\begin{algorithmic}[1]
\Require Episode frames $\{I_t\}_{t=1}^{N} \in \mathbb{R}^{N \times H \times W \times 3}$, joint trajectory $\{q_t\}$, gripper states $\{g_t\}$, open threshold $\tau_{\text{open}}$, initial extrinsic $T_{ec}^{0} \in SE(3)$
\State $\mathcal{F} \gets \{\, t : g_t < \tau_{\text{open}} \,\}$ \Comment{gripper-open frames}
\State $\{\tilde{I}_t\}_{t \in \mathcal{F}} \in \mathbb{R}^{T \times H' \times W'} \gets \textsc{GrayscaleAndCrop}(\{I_t\}_{t \in \mathcal{F}})$
\State $S_{\text{int}} \in \mathbb{R}^{H' \times W'} \gets \mathrm{median}_{t \in \mathcal{F}}\, \tilde{I}_t$ \Comment{pixelwise temporal median}
\State $S_{\text{std}} \in \mathbb{R}^{H' \times W'} \gets \mathrm{std}_{t \in \mathcal{F}}\, \tilde{I}_t$ \Comment{pixelwise temporal std}
\State $M_{\text{int}} \in \{0,1\}^{H' \times W'} \gets \mathbbm{1}\bigl[S_{\text{int}} < \mathrm{median}(S_{\text{int}})\bigr]$ \Comment{darkest half}
\State $M_{\text{std}} \in \{0,1\}^{H' \times W'} \gets \mathbbm{1}\bigl[S_{\text{std}} < \mathrm{median}(S_{\text{std}})\bigr]$ \Comment{most temporally rigid half}
\State $M_{\text{gt}} \in \{0,1\}^{H' \times W'} \gets M_{\text{int}} \cap M_{\text{std}}$
\State $\mathcal{L}(T) \gets 1 - \mathrm{IoU}\bigl(M(T; \{q_t\}),\, M_{\text{gt}}\bigr)$ \Comment{soft-IoU loss}
\State $T_{ec}^{*} \in SE(3) \gets \textsc{NelderMead}(\mathcal{L},\, T_{ec}^{0})$ \Comment{zeroth-order optimization}
\State \Return $T_{ec}^{*}$
\end{algorithmic}
\end{algorithm}

\section{Training Details}
\label{sec:supp_training}

\subsection{Mask Augmentation Details}
\label{subsec:supp_augmentation}

We detail here our three-stage mask augmentation procedure; each stage operates on the output of the previous one. Algorithm~\ref{alg:supp_augmentation} gives the full procedure.

\paragraph{Stage 1: Image rolling.} We first fill the masked region with surrounding image content by rolling the image over the initial mask $M$, replacing the source-embodiment pixels with content that is in-distribution with the rest of the image. Although masked pixels are nominally excluded by the vision encoder, two leak paths remain: the ViT only masks a patch when the \emph{majority} of its pixels are masked, so pixels in patches with only a few masked pixels still reach the encoder; and Stage 3 below explicitly drops parts of the mask, potentially re-exposing the source embodiment. Filling the body with rolled image content closes both.

\paragraph{Stage 2: Capsule attachment.} We then anchor capsules of random size and orientation at the boundary of the rolled mask. This distorts the source-embodiment silhouette and exposes the model to a broader distribution of end-effector shapes. We use $r_{\min}, r_{\max} = 12, 16$ pixels for the capsule radius, $L_{\min}, L_{\max} = 32, 64$ pixels for the capsule length, and $|\theta_{\max}| = 30^\circ$ for the capsule tilt from the vertical image axis.

\paragraph{Stage 3: Disk removal.} Finally, we remove disks of random size from the augmented mask, with centers sampled uniformly inside it. This further distorts the silhouette and additionally exposes the VLA to patch positional embeddings it would not otherwise see. We use $\rho_{\min}, \rho_{\max} = 8, 16$ pixels for the disk radius.

\begin{algorithm}[t]
\caption{Mask augmentation (training only).}
\label{alg:supp_augmentation}
\begin{algorithmic}[1]
\Require Wrist image $I \in \mathbb{R}^{H \times W \times 3}$, gripper mask $M \in \{0,1\}^{H \times W}$

\Statex \rule{0pt}{15pt}\textit{Stage 1: Image rolling.}
\State $\Delta_h \sim \mathcal{U}[H/3,\, 2H/3]$,\; $\Delta_w \sim \mathcal{U}[W/3,\, 2W/3]$
\State $\tilde{I} \gets \textsc{RollAndFill}(I, M, \Delta_h, \Delta_w)$

\Statex  \rule{0pt}{15pt}\textit{Stage 2: Capsule attachment.}
\State $N^+ \sim \mathcal{U}\{1, 2\}$
\For{$i = 1, \ldots, N^+$}
    \State $p_i \sim \mathcal{U}(\partial M)$ \Comment{on boundary}
    \State $r_i \sim \mathcal{U}[r_{\min}, r_{\max}]$
    \State $L_i \sim \mathcal{U}[L_{\min}, L_{\max}]$
    \State $\theta_i \sim \mathcal{U}[-\theta_{\max}, \theta_{\max}]$
    \State $C^+_i \gets \textsc{Capsule}(p_i,\, \theta_i,\, L_i,\, r_i)$
\EndFor
\State $M' \gets M \,\cup\, \bigcup_{i=1}^{N^+} C^+_i$

\Statex \rule{0pt}{15pt}\textit{Stage 3: Disk removal.}
\State $N^- \sim \mathcal{U}\{1, 2\}$
\For{$j = 1, \ldots, N^-$}
    \State $c_j \sim \mathcal{U}(M')$ \Comment{inside current mask}
    \State $\rho_j \sim \mathcal{U}[\rho_{\min}, \rho_{\max}]$
    \State $D^-_j \gets \textsc{Disk}(c_j,\, \rho_j)$
\EndFor
\State $\tilde{M} \gets M' \,\setminus\, \bigcup_{j=1}^{N^-} D^-_j$
\State \Return $\tilde{I},\, \tilde{M}$
\end{algorithmic}
\end{algorithm}

\section{Experimental Details}
\label{supp_exp}

\subsection{Task Definitions}
\label{subsec:supp_task_definitions}

We evaluate on four tabletop manipulation tasks, each divided into an ordered sequence of stages. We report two metrics. \emph{Progression rate} is the highest stage completed divided by the total number of stages. \emph{Success rate} is the binary indicator that the final stage was completed. Each baseline is run for 48 trials (12 trials per task), with each trial having randomized objects, placements, prompts, etc., and trials being generated once and reused identically for each baseline. Task definitions, progression stages, objects, and the 12 language prompts per task are listed together in Table~\ref{tab:supp_tasks}. Example trial setups are depicted in Figure~\ref{fig:supp_tasks}.

For all tasks, each object is placed at one of ``Front-Left'', ``Front-Right'', ``Back-Left'', or ``Back-Right'', relative to the robot base. For the UMI gripper, we replace the blue cube with the small cube for trials in which that object appears. This is due to a hardware issue in which the gripper firmware crashes when it is unable to reach a target position, which happens with the larger blue cube.

\subsection{IK Details}
\label{subsec:supp_ik}

\paragraph{src2trg (output actions).} For each step of an action chunk we fit the target embodiment to the source gripper's two tip poses. The tasks are listed in the top half of Table~\ref{tab:supp_ik_tasks}: two fingertip frame tasks, an orientation regularizer that anchors the wrist to its initial orientation so the solver does not jump between null-space branches, plus damping and posture regularizers. The damping cost is heavier on the arm than on the fingers so the solver prefers moving fingers when both can satisfy a target, and the posture term biases only the arm so the fingers are free to close into the grasp.

\paragraph{trg2src (input state).} To obtain the policy's state input, we solve for a source-gripper configuration that places its two gripper sites at the target's measured fingertip positions. The tasks are listed in the bottom half of Table~\ref{tab:supp_ik_tasks}: two position-only frame tasks, a matching wrist orientation regularizer, and damping and posture regularizers. The source's gripper opening is computed analytically from the distance between the two target tip positions. The heavy damping on the gripper DOFs is a redundant safeguard against any residual motion.

\begin{table}[t]
\centering
\small
\setlength{\abovecaptionskip}{10pt}
\renewcommand{\arraystretch}{1.15}
\begin{tabular}{l l l c c}
\toprule
\textbf{Direction} & \textbf{Task Description} & \textbf{Mink type} & \textbf{pos cost} & \textbf{ori cost} \\
\midrule
\multirow{5}{*}{\textit{src2trg}}
  & Left fingertip                              & FrameTask   & $1.0$    & $0.05$  \\
  & Right fingertip                             & FrameTask   & $1.0$    & $0.05$  \\
  & Wrist orientation regularizer               & FrameTask   & $0.0$    & $0.1$   \\
  & Damping (per-DOF: $1.0$ arm, $0.1$ hand)    & DampingTask & ---      & ---     \\
  & Posture (per-DOF: $10^{-2}$ arm, $0$ hand)  & PostureTask & ---      & ---     \\
\midrule
\multirow{5}{*}{\textit{trg2src}}
  & Left fingertip              & FrameTask   & $1.0$    & $0.0$   \\
  & Right fingertip             & FrameTask   & $1.0$    & $0.0$   \\
  & Wrist orientation regularizer               & FrameTask   & $0.0$    & $0.1$   \\
  & Damping (per-DOF: $1.0$ arm, $10^{3}$ gripper) & DampingTask & ---   & ---     \\
  & Posture (scalar $10^{-2}$, all DOFs)        & PostureTask & ---      & ---     \\
\bottomrule
\end{tabular}
\caption{Mink~\cite{mink} IK tasks for the two retargeting directions. Costs are unitless weights in the weighted-least-squares objective. All FrameTasks use a Levenberg-Marquardt damping coefficient of $1.0$.}
\label{tab:supp_ik_tasks}
\end{table}

\subsection{Baseline Details}
\label{subsec:supp_baselines}

All methods are evaluated on the same 48 trial instances per embodiment, on the same hardware, and under the same time limit and early-termination rule (Section~\ref{sec:exp:setup}).

\paragraph{\baselinePiDroid and ablations.}
\baselinePiDroid is the publicly available \baselinePi checkpoint finetuned on DROID with the same open-source recipe~\cite{openpi} that \vlaname{} follows, and it is deployed through the same tip-pose retargeting and IK parameters (Sections~\ref{sec:method:retargeting} and~\ref{subsec:supp_ik}). The two policies therefore differ only in whether the end-effector is cloaked. \baselineNoAug, \baselineOverlap, and \baselinePiDroidNoWrist share the data, training procedure, and retargeting of \vlaname{}, differing only in how the mask is constructed or, for \baselinePiDroidNoWrist, in the absence of the wrist camera input.

\paragraph{\baselineOverlap.}
\baselineOverlap is equivalent to \vlaname{} except for how masks are computed during training. \baselineOverlap pre-computes the target embodiment masks offline rather than augmenting the source mask in the manner described in Section~\ref{subsec:supp_augmentation}. Concretely, it is equivalent to replacing the ``add capsules'' part of the augmentation with ``add target masks''. Subtracting disks from the masks still occurs. The target embodiment masks are linearly interpolated between open and closed positions of each target embodiment, conditioned on the gripper position at each timestep. Everything else about \baselineOverlap is the same as \vlaname{}, including the mask construction, filling body pixels, masking the attention layers, and tip-pose retargeting.

\paragraph{\baselineLAP{}.}
Our \baselineLAP{}~\cite{zha2026laplanguageactionpretrainingenables} baseline uses their official open-source code and checkpoints. Since \baselineLAP{} is cross-embodiment by design, it required no retraining or architectural changes; adapting it to a new end-effector amounts to measuring the tip offset relative to the original gripper. Their deployment code for embodiments other than the Robotiq gripper was not publicly available, so we made some implementation choices ourselves. \baselineLAP{} outputs a 6-DoF wrist pose action. Since the UMI gripper and Sharpa hand share the same Franka arm as the Robotiq gripper, we reuse the IK code from \baselineLAP{} with per-embodiment wrist offsets. For the YAM arm and gripper, we had to implement our own IK pipeline. We again use Mink~\cite{mink} with objectives similar to our own method (discussed in Section~\ref{subsec:supp_ik}). Specifically, we use a FrameTask for the wrist and Posture and Damping tasks as regularizers on the arm.

\section{Additional Results}
\label{sec:supp_results}

\begin{table}[t]
\centering
\small
\setlength{\tabcolsep}{6pt}
\renewcommand{\arraystretch}{1.2}
\resizebox{\textwidth}{!}{%
\begin{tabular}{l l ccccc}
\toprule
 & & \multicolumn{4}{c}{Success rate (\%) $\uparrow$} & \\
\cmidrule(lr){3-6}
Embodiment & Method & Pick and Place & Remove & Move & Fold or Unfold & Task Average \\
\midrule
 \multirow{6}{2cm}{\textbf{Original Gripper}} & \baselinePiDroid\ver{v0} & \textbf{\val{100.0}{0.0}} & \textbf{\val{100.0}{0.0}} & \val{50.0}{15.1} & \val{58.3}{14.9} & \textbf{\val{77.1}{6.1}} \\
  & \baselineLAP & \val{83.3}{11.2} & \val{75.0}{13.1} & \textbf{\val{75.0}{13.1}} & \val{\underline{66.7}}{14.2} & \val{\underline{75.0}}{6.3} \\
  & \baselineNoAug\ver{v3.0} & \val{\underline{91.7}}{8.3} & \val{\underline{91.7}}{8.3} & \val{\underline{58.3}}{14.9} & \val{41.7}{14.9} & \val{70.8}{6.6} \\
  & \baselineOverlap\ver{v3.2} & \val{66.7}{14.2} & \val{\underline{91.7}}{8.3} & \val{50.0}{15.1} & \val{50.0}{15.1} & \val{64.6}{7.0} \\
  & \textbf{\vlaname{} (ours)}\ver{v3.1} & \val{\underline{91.7}}{8.3} & \val{83.3}{11.2} & \val{50.0}{15.1} & \textbf{\val{75.0}{13.1}} & \val{\underline{75.0}}{6.3} \\
  
\midrule
 \multirow{6}{2cm}{\textbf{UMI Gripper}\\(Unseen)} & \baselinePiDroid\ver{v0} + TP retargeting & \val{\underline{83.3}}{11.2} & \val{50.0}{15.1} & \val{16.7}{11.2} & \val{0.0}{0.0} & \val{37.5}{7.1} \\
  & \baselineLAP & \val{41.7}{14.9} & \val{33.3}{14.2} & \val{25.0}{13.1} & \val{25.0}{13.1} & \val{31.2}{6.8} \\
  & \baselineNoAug\ver{v3.0} & \val{66.7}{14.2} & \textbf{\val{83.3}{11.2}} & \textbf{\val{50.0}{15.1}} & \val{\underline{50.0}}{15.1} & \val{62.5}{7.1} \\
  & \baselineOverlap\ver{v3.2} & \textbf{\val{91.7}{8.3}} & \val{\underline{75.0}}{13.1} & \textbf{\val{50.0}{15.1}} & \val{41.7}{14.9} & \val{\underline{64.6}}{7.0} \\
  & \textbf{\vlaname{} (ours)}\ver{v3.1} & \textbf{\val{91.7}{8.3}} & \textbf{\val{83.3}{11.2}} & \val{\underline{41.7}}{14.9} & \textbf{\val{58.3}{14.9}} & \textbf{\val{68.8}{6.8}} \\
  
\midrule
 \multirow{6}{2cm}{\textbf{YAM Arm\\and Gripper}\\(Unseen)} & \baselinePiDroid\ver{v0} + TP retargeting & \val{41.7}{14.9} & \val{\underline{66.7}}{14.2} & \val{16.7}{11.2} & \val{50.0}{15.1} & \val{43.8}{7.2} \\
  & \baselineLAP & \val{33.3}{14.2} & \val{41.7}{14.9} & \val{33.3}{14.2} & \val{8.3}{8.3} & \val{29.2}{6.6} \\
  & \baselineNoAug\ver{v3.0} & \textbf{\val{83.3}{11.2}} & \val{50.0}{15.1} & \textbf{\val{66.7}{14.2}} & \val{33.3}{14.2} & \val{\underline{58.3}}{7.2} \\
  & \baselineOverlap\ver{v3.2} & \val{41.7}{14.9} & \val{\underline{66.7}}{14.2} & \val{25.0}{13.1} & \textbf{\val{66.7}{14.2}} & \val{50.0}{7.3} \\
  & \textbf{\vlaname{} (ours)}\ver{v3.1} & \val{\underline{75.0}}{13.1} & \textbf{\val{91.7}{8.3}} & \val{\underline{58.3}}{14.9} & \val{\underline{58.3}}{14.9} & \textbf{\val{70.8}{6.6}} \\
  
\midrule
 \multirow{6}{2cm}{\textbf{Sharpa Hand}\\(Unseen)} & \baselinePiDroid\ver{v0} + TP retargeting & \val{16.7}{11.2} & \val{25.0}{13.1} & \val{8.3}{8.3} & \val{16.7}{11.2} & \val{16.7}{5.4} \\
  & \baselineLAP & \val{\underline{66.7}}{14.2} & \val{16.7}{11.2} & \val{25.0}{13.1} & \val{16.7}{11.2} & \val{31.2}{6.8} \\
  & \baselineNoAug\ver{v3.0} & \val{41.7}{14.9} & \val{50.0}{15.1} & \val{\underline{33.3}}{14.2} & \val{\underline{33.3}}{14.2} & \val{39.6}{7.1} \\
  & \baselineOverlap\ver{v3.2} & \val{58.3}{14.9} & \val{\underline{58.3}}{14.9} & \textbf{\val{50.0}{15.1}} & \val{\underline{33.3}}{14.2} & \val{\underline{50.0}}{7.3} \\
  & \textbf{\vlaname{} (ours)}\ver{3.1} & \textbf{\val{75.0}{13.1}} & \textbf{\val{75.0}{13.1}} & \textbf{\val{50.0}{15.1}} & \textbf{\val{50.0}{15.1}} & \textbf{\val{62.5}{7.1}} \\
  
\bottomrule
\end{tabular}%
}
\caption{Zero-shot transfer across end-effectors. Each cell reports task success rate (mean $\pm$ SEM, \%). Success is the binary indicator that the policy reached the final progression stage. For each embodiment, \textbf{bold} marks the best score in the column and \underline{underline} marks the second-best.}
\label{tab:results_success}
\end{table}

\paragraph{Success rate.}
We report success rate in Table~\ref{tab:results_success}, the fraction of trials in which the policy completed the final progression stage as defined in Section~\ref{subsec:supp_task_definitions}. The relative ordering of methods within each embodiment is largely preserved, but the gap between \vlaname{} and the baselines is sharper here than on progression rate, since success rate offers no partial credit for late-stage failures.

\paragraph{Qualitative Rollouts.}
Figures~\ref{fig:supp_rollout_sharpa},~\ref{fig:supp_rollout_umi}, and~\ref{fig:supp_rollout_yam} show keyframe rollouts on the three unseen embodiments, comparing \vlaname{} against the \baselinePiDroid{}~+~retargeting and \baselineLAP{} baselines. Each row is one policy, ordered with \vlaname{} on top, and each column is a keyframe ordered in time; within a cell the wrist view is stacked above the external view. Across all three embodiments, closing the vision gap rather than refining the action interface is the key to cross-embodiment transfer. Figure~\ref{fig:supp_rollout_ablation} ablates the masking recipe itself on the Sharpa hand.

\newpage

\begin{figure}[t]
    \centering
    \includegraphics[width=\linewidth]{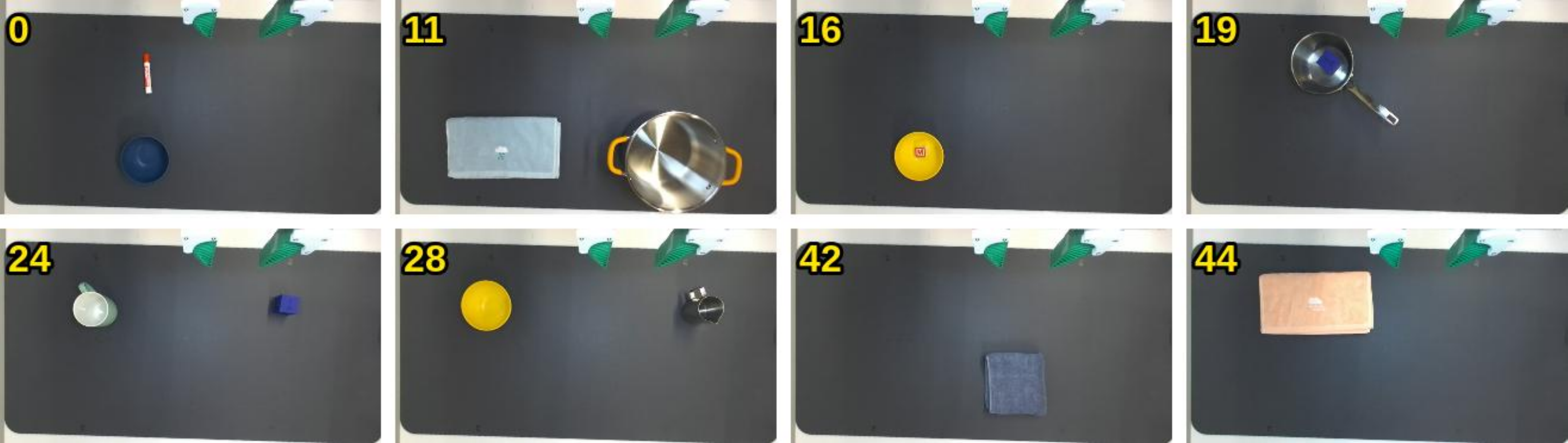}
    \vspace{-1.5em}
    \caption{Example trial setups across the four tasks (two per task). The bright yellow number overlaid on each image is the trial index, matching the corresponding row in Table~\ref{tab:supp_tasks}.}
    \label{fig:supp_tasks}
\end{figure}

\begin{table}[t]
\centering
\scriptsize
\setlength{\tabcolsep}{4pt}
\renewcommand{\arraystretch}{1.0}
\begin{tabular}{@{}p{0.18\linewidth}p{0.22\linewidth}p{0.13\linewidth}p{0.08\linewidth}p{0.32\linewidth}@{}}
\toprule
\textbf{Task \& description} & \textbf{Progression stages} & \textbf{Objects} & \textbf{Trial \newline index} & \textbf{Language prompt} \\
\midrule
\textbf{Pick and Place.} \newline Pick an object with a stable grasp and place it in a container.
& (1) Reaches the correct object \newline
  (2) Firmly grasps and lifts without dropping \newline
  (3) Reaches the container with object in hand \newline
  (4) Places the object in the container
& Red marker \newline
  Black marker \newline
  Small cube \newline
  Blue cube \newline
  Grey towel \newline
  Pink towel \newline
  Blue towel \newline
  Blue bowl \newline
  Orange bowl \newline
  Small pot \newline
  Large pot
& 0 \newline 1 \newline 2 \newline 3 \newline 4 \newline 5 \newline 6 \newline 7 \newline 8 \newline 9 \newline 10 \newline 11
& ``Place the marker in the blue bowl'' \newline
  ``Put the pen in the bowl'' \newline
  ``Place the red marker in the orange bowl'' \newline
  ``Place the small block in the bowl'' \newline
  ``Place the block in the bowl'' \newline
  ``Place the small block in the blue bowl'' \newline
  ``Put the blue block in the small pot'' \newline
  ``Place the small cube in the small pot'' \newline
  ``Put the small cube in the small pot'' \newline
  ``Place the grey towel in the large pot'' \newline
  ``Put the towel in the large pot'' \newline
  ``Put the blue cloth in the large pot'' \\
\midrule
\textbf{Remove.} \newline Remove an object from inside a container and place it on the table.
& (1) Reaches the container \newline
  (2) Firmly grasps the object inside and lifts it clear \newline
  (3) Places the object on the table
& Red marker \newline
  Blue marker \newline
  Small cube \newline
  Blue cube \newline
  Grey towel \newline
  Pink towel \newline
  Blue towel \newline
  Blue bowl \newline
  Yellow bowl \newline
  Orange bowl \newline
  Small pot
& 12 \newline 13 \newline 14 \newline 15 \newline 16 \newline 17 \newline 18 \newline 19 \newline 20 \newline 21 \newline 22 \newline 23
& ``Remove the marker from the bowl'' \newline
  ``Remove the marker from the yellow bowl'' \newline
  ``Take the blue pen out of the orange bowl'' \newline
  ``Remove the blue cube from the blue bowl'' \newline
  ``Take the small cube out of the yellow bowl'' \newline
  ``Take the small cube out of the blue bowl'' \newline
  ``Take the block out of the small pot'' \newline
  ``Take the blue cube out of the small pot'' \newline
  ``Remove the small cube from the small pot'' \newline
  ``Take the grey towel out of the small pot'' \newline
  ``Remove the cloth from the small pot'' \newline
  ``Remove the blue cloth from the small pot'' \\
\midrule
\textbf{Move.} \newline Transport an object to a position next to a target object (not inside it).
& (1) Reaches the object \newline
  (2) Firmly grasps and lifts without dropping \newline
  (3) Reaches the target with object in hand \newline
  (4) Places the object next to the target
& Small cube \newline
  Blue cube \newline
  Silver cup \newline
  White cup \newline
  Green cup \newline
  Grey towel \newline
  Pink towel \newline
  Blue towel \newline
  Yellow bowl \newline
  Orange bowl \newline
  Blue bowl \newline
  Small pot
& 24 \newline 25 \newline 26 \newline 27 \newline 28 \newline 29 \newline 30 \newline 31 \newline 32 \newline 33 \newline 34 \newline 35
& ``Move the cube next to the cup'' \newline
  ``Move the cube next to the white mug'' \newline
  ``Move the block next to the cup'' \newline
  ``Move the blue block near the white mug'' \newline
  ``Move the cup near the yellow bowl'' \newline
  ``Move the silver mug near the bowl'' \newline
  ``Move the white mug near the bowl'' \newline
  ``Move the mug next to the bowl'' \newline
  ``Move the pink towel next to the small pot'' \newline
  ``Move the grey towel near the small pot'' \newline
  ``Move the towel near the small pot'' \newline
  ``Move the blue towel next to the small pot'' \\
\midrule
\textbf{Fold / Unfold.} \newline Apply one additional fold to a pre-folded towel, or unfold a pre-folded towel by one step. Folded result should be mostly folded and not too untidy; unfolded result should be mostly flat and uncrumpled.
& (1) Reaches the towel \newline
  (2) Firmly grasps and manipulates it without slipping \newline
  (3) Completes the fold (or unfold) and releases
& Grey towel \newline
  Pink towel \newline
  Blue towel
& 36 \newline 37 \newline 38 \newline 39 \newline 40 \newline 41 \newline 42 \newline 43 \newline 44 \newline 45 \newline 46 \newline 47
& ``Unfold the towel'' \newline
  ``Unfold the towel'' \newline
  ``Unfold the towel'' \newline
  ``Unfold the towel'' \newline
  ``Unfold the cloth'' \newline
  ``Fold the towel'' \newline
  ``Fold the cloth'' \newline
  ``Unfold the cloth'' \newline
  ``Unfold the towel'' \newline
  ``Unfold the towel'' \newline
  ``Fold the grey cloth'' \newline
  ``Unfold the pink cloth'' \\
\bottomrule
\end{tabular}
\caption{Task definitions, progression stages, objects, and the per-trial language prompts.}
\label{tab:supp_tasks}
\end{table}

\clearpage
\begin{figure}[t]
    \centering
    \includegraphics[width=\columnwidth]{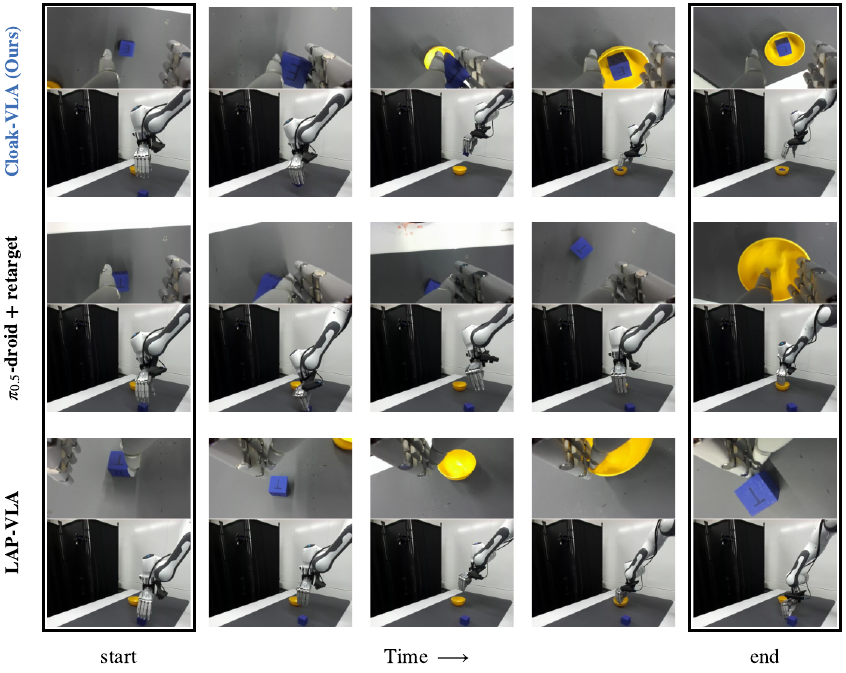}
    \caption{Keyframe rollouts on the unseen Sharpa hand for the pick-and-place task, prompt \textit{``Place the block in the bowl''}. \baselinePiDroid{}~+~retargeting never hides the body, so its wrist view stays out of distribution and it grasps inaccurately and aims off target. \baselineLAP{} aligns the embodiments at the language layer, giving a cleaner action interface, but cannot close the vision gap: it fails to grasp and arrives at the bowl empty-handed. \vlaname{} grasps and places cleanly.}
    \label{fig:supp_rollout_sharpa}
\end{figure}

\begin{figure}[t]
    \centering
    \includegraphics[width=\columnwidth]{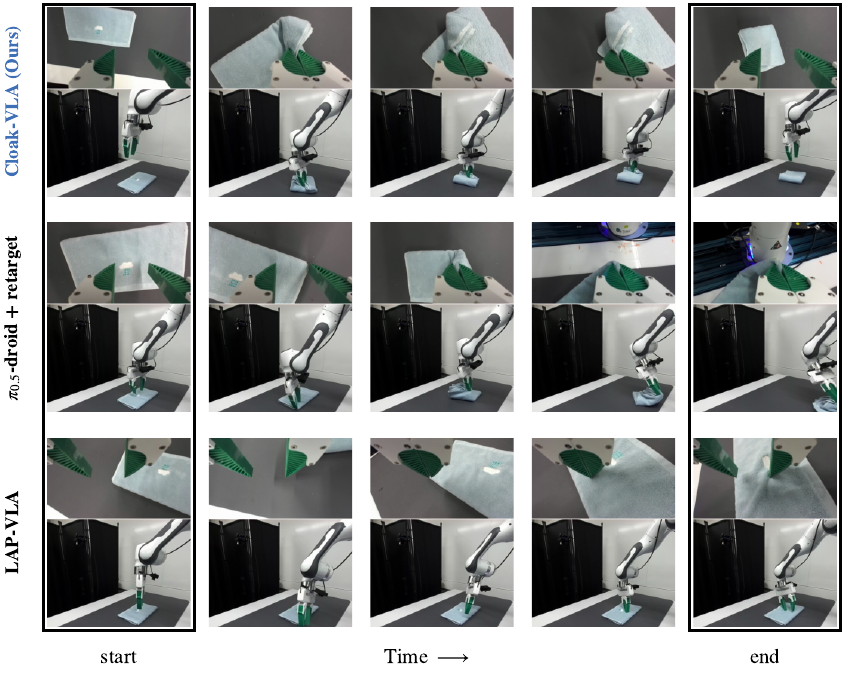}
    \caption{Keyframe rollouts on the unseen UMI gripper for the fold task, prompt \textit{``Fold the towel''}. \baselinePiDroid{}~+~retargeting grasps and lifts the towel, but with the body still in the wrist view its actions drift out of distribution and the fold collapses. \baselineLAP{} keeps a clean action interface, yet the vision gap throws off its aim and it never lines up to grasp. \vlaname{} folds the towel cleanly.}
    \label{fig:supp_rollout_umi}
\end{figure}

\begin{figure}[t]
    \centering
    \includegraphics[width=\columnwidth]{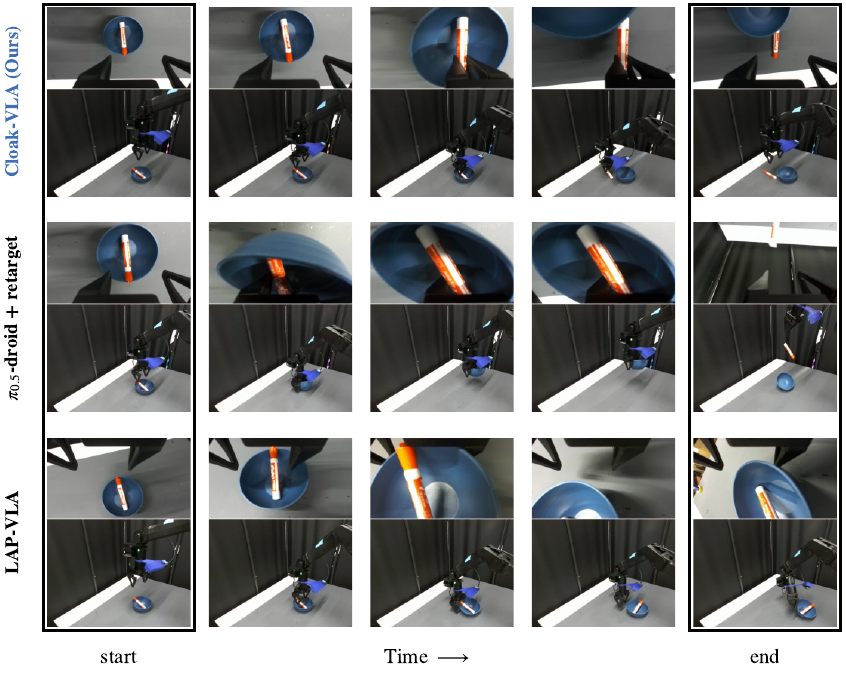}
    \caption{Keyframe rollouts on the unseen YAM arm and gripper for the remove task, prompt \textit{``Remove the marker from the bowl''}. Unlike the UMI and Sharpa embodiments, which keep the source Franka arm and change only the end-effector, the YAM is a different arm \emph{and} gripper. \baselinePiDroid{}~+~retargeting grasps the marker and the bowl together and drifts out of distribution; \baselineLAP{} grasps at empty air. \vlaname{} removes the marker cleanly, showing that cloaking the wrist view is enough to cross arms, not just end-effectors.}
    \label{fig:supp_rollout_yam}
\end{figure}

\begin{figure}[t]
    \centering
    \includegraphics[width=\columnwidth]{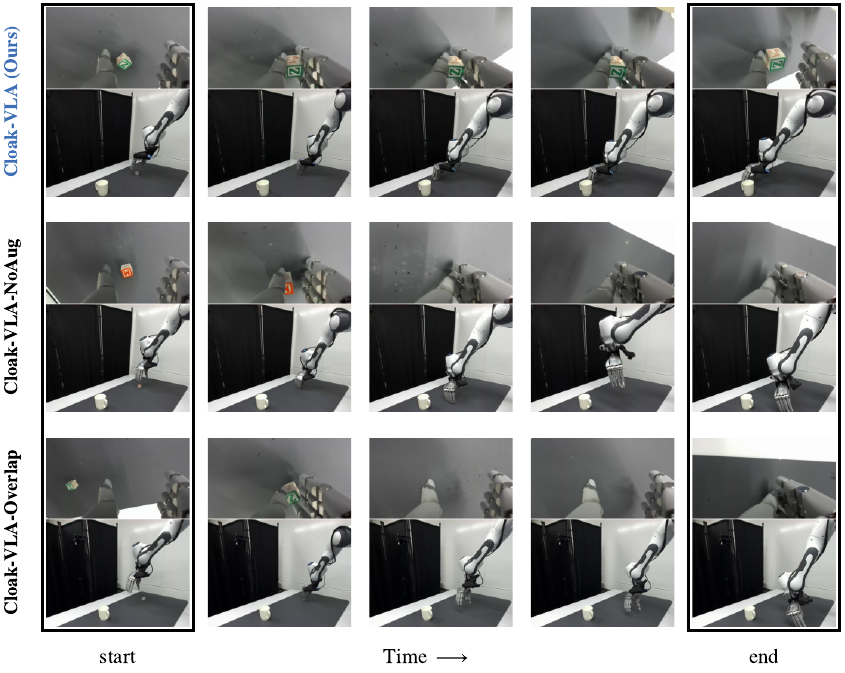}
    \caption{Ablation rollouts on the unseen Sharpa hand for the move task, prompt \textit{``Move the cube next to the white mug''}. We compare \vlaname{} against two ablations that share its architecture but change the training mask: \baselineNoAug{}, trained on the bare source-embodiment mask with no augmentation, and \baselineOverlap{}, a Shadow-inspired \citep{lepert2024shadow} variant that overlays the source mask with a known target silhouette. Both ablations either omit augmentation or assume the target embodiment is known in advance, isolating mask augmentation as the component that closes the embodiment gap.
    Only \vlaname{} grasps the cube and completes the move, while both ablations fail to secure it, highlighting that mask augmentation is necessary for a robust wrist-view representation on an unseen embodiment.}
    \label{fig:supp_rollout_ablation}
\end{figure}

\clearpage
\bibliography{example}  %

\else

\begin{abstract}
We present \methodname{}, a training recipe that endows a Vision-Language-Action (VLA) model with zero-shot cross-embodiment transfer by cloaking the end-effector from its own wrist camera. The end-effector occupies a large and consistent region of the wrist view and masking it allows for embodiment-agnostic visual reasoning. \methodname{} renders a mask in simulation from the robot's known geometry, accurately and in real time, with no segmentation or generative models. During training, we augment the mask so the model generalizes to embodiments unseen at training time. We demonstrate the recipe with \vlaname{}, a VLA trained with \methodname{} on a single parallel-jaw gripper dataset. No data of new embodiments is ever collected. \vlaname{} transfers zero-shot to various unseen embodiments, including another gripper, another arm, and a five-fingered hand, while preserving the source embodiment's performance. By decoupling the wrist view from its own embodiment, \methodname{} allows data to outlive the hardware it was collected on.

\end{abstract}

\keywords{Cross-Embodiment Transfer, Vision-Language-Action Models, Manipulation}

\begin{figure}[H]
  \centering
  \includegraphics[width=\linewidth]{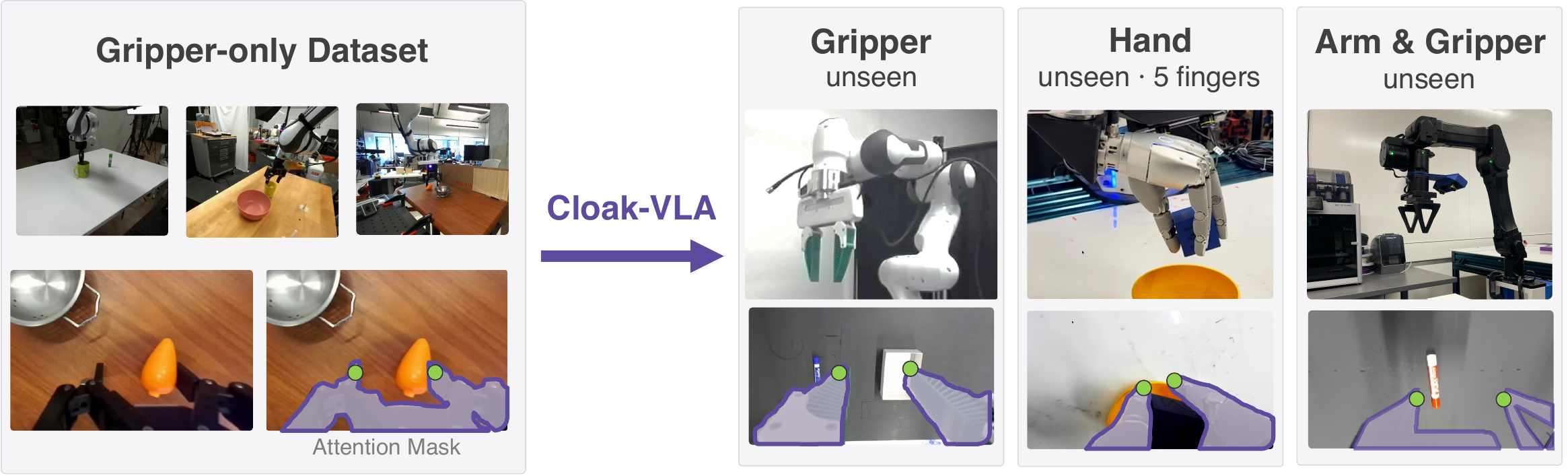}
  \vspace{-5pt}
\caption{
We cloak the end-effector from its own wrist camera, letting a VLA trained on a single parallel-jaw gripper transfer zero-shot to unseen embodiments, including a five-fingered hand.
}
\label{fig:teaser}
\end{figure}

\section{Introduction}
\label{sec:intro}

Robotic manipulation is fragmented across a diverse landscape of end-effectors, from parallel-jaw grippers to increasingly multi-fingered, anthropomorphic hands.
Collecting demonstrations on a specific embodiment (the source) and training a model on them yields strong performance on that embodiment, but the returns diminish as the deployment embodiment (the target) diverges from the source. Collecting new data for each target embodiment is possible, but costly, especially in a rapidly evolving hardware landscape.
The tension is already concrete; the community has amassed enormous gripper-centric datasets~\citep{open_x_embodiment_rt_x_2023,khazatsky2024droid,chi2024universal} while the frontier broadens to dexterous and humanoid hands~\citep{gr00tn1_2025,xu2025dexumi}, raising the question of how gripper-centric data can be repurposed for new end-effectors.
Our goal is to train a single policy deployable on target embodiments never seen during training, without collecting data on those new embodiments, i.e. zero-shot cross-embodiment transfer.

Reaching zero-shot transfer on a new embodiment is difficult due to the visual and kinematic gap between the source and target embodiments. The policy must drive unfamiliar joints, conform to different link geometries, and handle different end-effector appearances and camera angles.
The kinematic mismatch has a structured, geometric character, which can be overcome with inverse kinematics. The visual gap, that is, the end-effector's appearance in its own wrist camera, is less straightforward to close, and is the binding constraint on zero-shot transfer.

A perception stack trained on a single embodiment bakes its appearance into its features. The policy becomes accustomed to the source embodiment, therefore even a slightly different target embodiment pushes the visual observation out of distribution and degrades performance. Unlike a mismatch in states or actions, this failure cannot be repaired by retargeting the representation.
This gap is most severe in the common setup with wrist-mounted cameras~\cite{chi2024universal,khazatsky2024droid}, where the end-effector fills a large and consistent portion of every frame. This problem is only exacerbated as the field shifts from compact grippers to articulated, multi-finger hands that bear ever less resemblance to a gripper source embodiment.
One line of work, therefore, attempts to overwrite the visual observation by repainting or regenerating the image with generative models~\citep{chen2024roviaug,yuan2025roboengine}. But such models are prone to hallucination and ill-suited to real-time control.

In this work we present \methodname{} (Figure~\ref{fig:teaser}), a recipe for making a Vision-Language-Action (VLA) policy transfer zero-shot to embodiments not seen during training by cloaking the end-effector from its own wrist view.
We demonstrate \methodname{}'s effectiveness by training \vlaname{}, a VLA finetuned from $\pi_{0.5}$~\citep{black2025pi_} on a single parallel-jaw gripper source embodiment, that transfers zero-shot to different target embodiments, including a five-fingered hand.
During training, we mask image patches in the visual encoder that are occupied by the gripper, cloaking it from the policy's perception.
This mask is obtained directly from the robot's proprioceptive state and the wrist camera parameters using a simple forward kinematics and projection operation. We further augment the mask during training to prevent overfitting to the source gripper's silhouette. With no reliance on generative models to modify visual observations, our approach is immune to model hallucination.

Deployment is then straightforward; we mask the target end-effector the same way, without mask augmentations, and retarget the proprioceptive state and actions through the two fingertips using inverse kinematics. By doing this, the visual and proprioceptive observations both appear in-distribution to the policy, sidestepping the need to collect data and retrain on the new embodiment. Because the focus of our work is addressing observation gaps, not action gaps, we employ a simple manipulation skill reachable by placing two tips, whether gripper jaws or fingertips, rather than morphology-specific dexterous skills that require new contact strategies.

We validate \vlaname{} on three embodiments not seen during training: a parallel-jaw gripper and a five-fingered hand, both mounted on the same arm as the source, together with an additional arm platform.
We first measure the cost of \methodname{} to the source embodiment, where there is no transfer gap, and find only a small performance drop relative to the unmasked policy.
More strikingly, on the unseen embodiments \vlaname{} still retains most of the original performance, and far outpaces every baseline. This is most clear on the five-fingered hand, where the unmasked policy degrades sharply while \vlaname{} retains strong performance.
We view \methodname{} as a step toward decoupling manipulation skill from the body that executes it so that data gathered on one embodiment can outlive the hardware it was collected on.

\section{Related Work}

\subsection{Cross-embodiment generalist policies}
Most cross-embodiment policies combine two practices: scaling data across many robots and adding a representation that absorbs their heterogeneity.
Open X-Embodiment's coarse end-effector alignment~\citep{open_x_embodiment_rt_x_2023} and generalist policies like OpenVLA~\citep{kim24openvla} and $\pi_0$~\citep{black2024pi_0,black2025pi_} are examples of combining demonstrations from many robot embodiments.
Explicit mechanisms for learning cross-embodiment representations include identity embeddings or soft prompts in GR00T~\citep{gr00tn1_2025} and X-VLA~\citep{zheng2025x}, per-embodiment tokenizers in Octo~\citep{octo_2023}, CrossFormer~\citep{Doshi24-crossformer}, and HPT~\citep{wang2024hpt}, or padded action vectors with per-robot slots in RDT-1B~\citep{liu2025rdt} and $\pi_0$~\citep{black2024pi_0}.
Most methods leverage both, yet the policy is fundamentally tied to the training embodiments. Supporting a new robot requires collecting new training data or adding another learned module.

Recent works claim true zero-shot transfer to unseen embodiments, yet the leading examples share a common limitation.
RDT2~\citep{liu2026rdt2} trains on thousands of hours of UMI gripper data and transfers across robot arms, but always with the same parallel-jaw gripper it was trained on.
LAP~\citep{zha2026laplanguageactionpretrainingenables} casts actions as language and transfers even across grippers, yet these are invariably parallel-jaw with the same color, such that the end-effector's appearance hardly changes.
In both RDT2 and LAP, the end-effector region of the wrist view looks essentially the same as in training. Neither confronts the visual gap, which is the decisive factor governing cross-embodiment transfer in VLAs.
By masking that gap directly through the vision encoder, \methodname{} transfers from a single source gripper, with no conditioning or target data, to unseen embodiments.

\subsection{Closing the visual embodiment gap}
A complementary line of work addresses the visual gap directly by editing the appearance of the robot in the image.
One strategy repaints a replacement embodiment. Mirage~\citep{chen2024mirage} uses known models and camera parameters to mask out the unseen target robot, inpaint the hole, and overlay a render of the source robot at test time.
A parallel line of work inpaints human video, editing the person out and rendering a robot in their place~\citep{bahl2022human,dessalene2025embodiswap,lepert2025phantom,lepert2025masquerade,li2025h2r}, while others synthesize alternative robots~\citep{chen2024roviaug,ji2025oxeauge} or randomize appearance in simulation~\citep{dan2025x}.
In all of these approaches, the replacement embodiment must be blended into the scene; renders can clash with the real background, and generative variants are heavy, unrealistic, and prone to hallucination.

Rather than replacing the robot view, another approach removes it entirely.
Shadow~\cite{lepert2024shadow} overlays a composite silhouette of the source and target robots during training.
Crucially, the target embodiment must be known at training time. The transfer is zero-shot only to a robot fixed in advance, not to any future one, and only when source and target are similar enough.
ARRO~\citep{mirjalili2025augmented} and RoboEngine~\cite{yuan2025roboengine}
instead use learned segmentation. EgoMimic~\citep{kareer2024egomimic} masks the human hand and robot arm to align human and robot video, but co-trains on both rather than transferring to an unseen embodiment. But learned masks are unreliable, mislabeling objects as part of the robot, leaving it partly unmasked, and degrading on thin structures.
\methodname{} instead masks the embodiment from its known geometry, with nothing to render or hallucinate, and augments the mask during training such that the target need not be known in advance, transferring zero-shot to unseen embodiments, including a five-fingered hand.

\subsection{Cross-embodiment dexterous manipulation}
Cross-embodiment research on dexterous hands has concentrated on control and morphology.
Fingertip and keypoint retargeting maps a common target onto many hands and underpins teleoperation across embodiments, as in DexPilot~\citep{handa2020dexpilot} and AnyTeleop~\citep{qin2023anyteleop}.
Building on this, one line of work learns a single policy that transfers across morphologies through a shared action space: eigengrasps in CrossDex~\citep{yuan2025cross}, a canonical URDF in One-Hand-to-Rule-Them-All~\citep{wei2026one}, shared action latents~\citep{bauerlatent}, and morphology inferred on the fly in DexFormer~\cite{dexformer2026}.
DexFormer and One-Policy-Fits-All~\citep{mu2026one} learn a single policy that transfers zero-shot across hands, closing the morphology and control gap while leaving the visual gap untouched.
DexUMI~\citep{xu2025dexumi} confronts the dexterous visual gap by removing the human hand from wrist video and compositing the real robot hand back in, but does so for data collection and requires segmentation, inpainting, and a physical replay of each target hand.
Therefore, transferring to a new multi-finger hand zero-shot without collecting new data remains open.

\section{Preliminaries}
\label{sec:prelim}

\textbf{A Vision-Language-Action (VLA) model} is a policy that maps one or more camera images, a language instruction, and the robot's proprioceptive state to robot actions, typically by attaching an action head to a pretrained vision-language model and training via imitation learning on teleoperated demonstrations~\citep{kim24openvla,octo_2023,black2024pi_0}.
Our method targets two recurring structural components of modern VLAs: a vision transformer (ViT) \emph{vision encoder} that tokenizes each camera image into patches, and an \emph{action head} that decodes a chunk of future actions from those image tokens together with the language and the state. \methodname{} computes an attention mask on the vision encoder patches from the geometric embodiment mask.

\baselinePi~\citep{black2025pi_}, our base policy, is a recent VLA built on $\pi_0$~\citep{black2024pi_0}. In $\pi_{0.5}$, wrist camera images are tokenized into patches by a ViT-style vision encoder, fused with the language prompt and the proprioceptive state inside the VLA backbone, and decoded into an action chunk by the action head. The proprioceptive state input to \baselinePi is the robot's joint positions, and the output is joint position actions in the same space. We finetune \baselinePi on the DROID dataset~\cite{khazatsky2024droid}, so that the proprioceptive input state and output actions are in the 8 DoF joint space (7 DoF arm and 1 DoF gripper) of the DROID data's source embodiment. The visual input is a wrist camera view and an external camera view.

\section{Method}
\label{sec:method}

\begin{figure}[t]
  \centering
  \includegraphics[width=\linewidth]{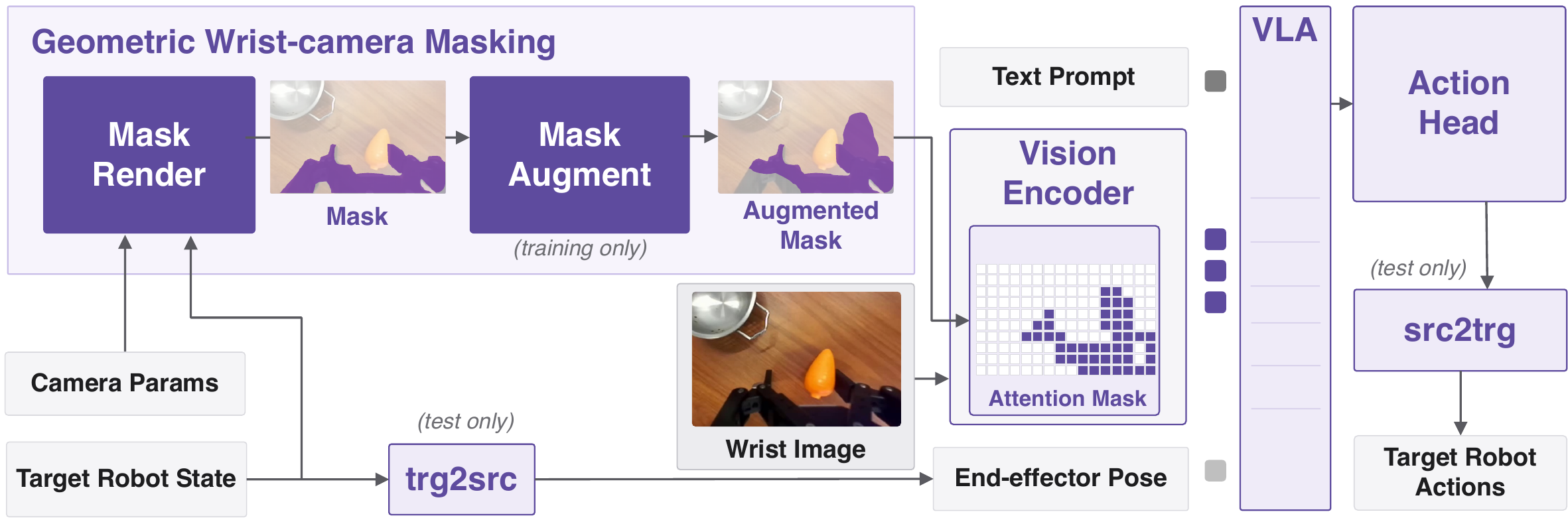}
  \vspace{-10pt}
\caption{\textbf{Overview.}
\methodname{} renders a geometric mask of the end-effector using the robot state and wrist camera parameters, augments it during training, and uses it to compute an attention mask for the vision encoder, cloaking the end-effector from the wrist view.
The resulting image patch tokens, the source-robot state, and the language prompt drive a single VLA backbone and action head.
On an unseen embodiment, tip-pose retargeting bridges the robots via FK and IK; \emph{trg2src} maps the target state to the source state input to the policy; \emph{src2trg} maps action outputs back to the target.
}
\vspace{-10pt}
  \label{fig:overview}
\end{figure}

In this section, we describe our recipe for learning a VLA with zero-shot cross-embodiment transfer. Figure~\ref{fig:overview} shows an overview of our method.
Our goal is to train a policy on a single source embodiment that deploys zero-shot on unseen target embodiments with no target data, no retraining, and no need to know the target in advance.
To address the visual gap, \methodname{} computes an embodiment mask in real time, rasterizing the end-effector meshes in simulation from the robot's description file, its proprioceptive state, and the wrist camera's parameters.
During training, we randomize the mask shape so the policy never sees a fixed gripper silhouette (\S~\ref{sec:method:mask}).
The mask is then downsampled to the vision encoder's patch grid resolution so that patches aligned with the mask are removed from the attention computation (\S~\ref{sec:method:attention}).
To address the kinematic gap, we leverage forward kinematics (FK) and inverse kinematics (IK). The state input and action output of the VLA stay in the native source embodiment's space, untouched at training time.
To deploy on an unseen end-effector, we apply \emph{tip-pose retargeting} (\S~\ref{sec:method:retargeting}), converting state and actions between source and target embodiments. Given the action output of the VLA, the 6-DoF poses of two designated source embodiment tips are computed via FK. The target embodiment's actions are then computed via IK on corresponding tips. A similar operation is done to compute the source embodiment state input to the VLA.
With the \methodname{} recipe in hand, we finetune $\pi_{0.5}$ on a single parallel-jaw gripper dataset (\S~\ref{sec:method:training}), yielding \vlaname{}, our trained zero-shot cross-embodiment policy.

\subsection{Embodiment mask}
\label{sec:method:mask}

\paragraph{Construction.}
Let $I \in \mathbb{R}^{H \times W \times 3}$ be the wrist-camera image, $q$ the robot's proprioceptive state, and $\mathcal{G}$ the set of link meshes given by the robot description file. The wrist camera's pose in the end-effector frame, $T_{ec} \in SE(3)$, is obtained per episode via \emph{Silhouette Calibration} (Section~\ref{subsec:supp_extrinsics}, Algorithm~\ref{alg:supp_extrinsics}). Composed with the end-effector's world pose from forward kinematics, $T_{we}(q) = \mathrm{FK}_{\text{ee}}(q)$, this gives the camera's world pose $ T_{wc}(q) = T_{we}(q)\,T_{ec}$. We pose each link mesh into the world via forward kinematics: for each link $\ell \in \mathcal{G}$ with vertices $V_\ell$, $T_{w\ell}(q) = \mathrm{FK}_\ell(q)$ gives its world pose. Projecting the union into the camera with intrinsics $K$ and rasterizing yields the embodiment mask, as shown in Equation~\ref{eq:rasterize}. Additionally, we dilate $M$ by a small morphological kernel to absorb residual calibration error.

\begin{equation}
  M \;=\; \mathrm{rasterize}\!\left(\,
    \bigcup_{\ell \in \mathcal{G}}
    K\, T_{wc}(q)^{-1}\, T_{w\ell}(q)\, V_\ell
  \,\right) \;\in\; \{0,1\}^{H \times W}
  \label{eq:rasterize}
\end{equation}

\paragraph{Augmentation.}
During training only, we augment the mask shape so the policy sees a range of plausible end-effector silhouettes. Specifically, to distort the source embodiment's silhouette, we union $M$ with a random set of capsules $C^+_i$ anchored on its contour and subtract a random set of disks $D^-_j$ from its interior, as shown in Equation~\ref{eq:augmentation}. Capsule and disk counts, radii, lengths, and placements are detailed in Section~\ref{subsec:supp_augmentation}.
\begin{equation}
  \tilde{M} \;=\; \Bigl(\,M \;\cup\; \bigcup_{i=1}^{N^+} C^+_i\,\Bigr) \;\setminus\; \bigcup_{j=1}^{N^-} D^-_j
  \label{eq:augmentation}
\end{equation}

\paragraph{Filling body pixels.}
Because our augmentation procedure subtracts from the mask, it can reveal the underlying source embodiment. To avoid this, we fill in image content onto $M$. Rather than inpaint with a solid color or random noise, which would not match the pixel distribution at inference time, or with a generative model, which is overkill, we simply inpaint a rolled copy of the same image onto $M$:
\begin{equation}
  \tilde{I}(h, w) \;=\;
  \begin{cases}
    I\bigl((h + \Delta_h) \bmod H,\;(w + \Delta_w) \bmod W\bigr) & (h, w) \in M, \\
    I(h, w) & \text{otherwise},
  \end{cases}
  \label{eq:image_roll}
\end{equation}
with $\Delta_h, \Delta_w \sim \mathcal{U}[H/3,\,2H/3]$. At test time we drop the augmentations and apply only the construction step. See Section~\ref{subsec:supp_augmentation} for more discussion.

\subsection{Masked attention in the vision encoder}
\label{sec:method:attention}

The augmented wrist image $\tilde I$ enters the visual encoder, a ViT-style transformer (the $\pi_{0.5}$ encoder in our experiments), which splits it into patches of size $P \times P$ pixels. The image-resolution mask $\tilde M$ is downsampled to the patch grid by majority vote per patch (a patch is marked masked when more than half of its pixels are masked), and flattened, yielding $\tilde{M}^{\mathcal{A}} \in \{0,1\}^{H_p \cdot W_p}$ with $H_p = H/P$, $W_p = W/P$. For the \baselinePi vision encoder, $P=14$ and $H=W=224$, so $\tilde{M}^{\mathcal{A}} \in \{0, 1\}^{256}$.

We keep all patch tokens in the sequence but bias the attention so no query reads from a masked patch. In every self-attention layer, with queries, keys, and values $Q, K, V \in \mathbb{R}^{N \times d}$.

\begin{equation}
  \mathrm{Attn}(Q, K, V) \;=\; \mathrm{softmax}\!\left(\frac{Q K^\top}{\sqrt{d}} \;+\; A\right) V
  \qquad
  A_{ij} = \begin{cases} -\infty & \text{if } \tilde{M}^{\mathcal{A}}_j = 1, \\ 0 & \text{otherwise}. \end{cases}
\end{equation}

Masked tokens keep their sequence slots but contribute nothing to other tokens, equivalent to dropping them at constant sequence length. The same $\tilde{M}^{\mathcal{A}}$ is reused across all self-attention layers and propagated to the VLA backbone, where it gates masked patch tokens out of the downstream cross-modal attention with the language and state. We find that the external camera view contributes little to the visual embodiment gap, and therefore do not mask it.

\subsection{Tip-pose retargeting}
\label{sec:method:retargeting}

\vlaname{}'s inputs and outputs are in the source robot's joint space, so deploying on an unseen target requires converting between two embodiments. Even if the embodiments share the same arm, the geometry of the end-effector may require different arm configurations to accomplish the same task, again necessitating retargeting. We do this through end-effector tips, designating two tips on each embodiment to bridge their action spaces and using FK on the model output followed by IK to achieve the target embodiment joint actions. To compute the model input in the next step, we run FK on the target embodiment then IK on the source embodiment. Matching tips between grippers is straightforward. For the five-fingered hand, we match the left and right parallel jaw gripper tips (in the wrist image) of our source embodiment to the thumb and middle fingertips, respectively. 

We use $\hat{\cdot}$ to indicate an action quantity from the model output as opposed to a state quantity for the model input. Given a source joint action output by the model, $\hat{q}_{src}$, we run FK to compute the poses of the source tips, $(\hat{T_{l}}, \hat{T_{r}}) = \mathrm{FK}(\hat{q}_{src})$. $\hat{T_{l}}, \hat{T_{r}} \in \mathrm{SE}(3)$ define the objective function for the IK step: $\hat{q}_{trg} = \mathrm{IK}(\hat{T_{l}}, \hat{T_{r}})$. This is done for every action in the action chunk. To compute the model input state $q_{src}$, we go through the two tips again: $T_l, T_r = \mathrm{FK}(q_{trg})$ and $q_{src} = \mathrm{IK}(T_l, T_r)$.

In practice, the $\mathrm{IK}(\mathrm{FK}(\cdot))$ operation is used to position the robot arm, while the end-effector solution to the IK is overwritten by the gripper signal output by the model, $\hat{g} \in \mathbb{R}$, which is part of $\hat{q}_{src}$. For the hand target embodiment, we use $\hat{g}$ to linearly interpolate between predefined open and closed hand configurations. We also include regularizing terms in the IK objective to avoid instability in null-space solutions. See Section~\ref{subsec:supp_ik} for a detailed explanation of our IK implementation, which uses Mink~\cite{mink}.

\subsection{Training}
\label{sec:method:training}

We start from a single-embodiment dataset of tuples $(I, I_{\mathrm{ext}}, q, \ell, a)$, where $I$ is the wrist image, $I_{\mathrm{ext}}$ is the external view, $q$ is the proprioceptive state, $\ell$ is the language, and $a$ is the output action chunk, all in the source robot's native joint space. The source embodiment mask $M$ (\S~\ref{sec:method:mask}) is computed offline, yielding modified samples $(I, I_{\mathrm{ext}}, M, q, \ell, a)$.
At each training step we augment $M \to \tilde M$ and fill body pixels $I \to \tilde I$ online (\S~\ref{sec:method:mask}), then downsample to the patch-grid mask $\tilde{M}^{\mathcal{A}}$ consumed by the encoder (\S~\ref{sec:method:attention}).
\vlaname{} is obtained by finetuning a pre-trained $\pi_{0.5}$ checkpoint on this modified dataset, following a standard open-source finetuning procedure~\cite{openpi}.

\section{Experiments}
\label{sec:experiments}

\methodname{} targets zero-shot cross-embodiment manipulation by cloaking the end-effector from its own wrist camera such that a policy trained on a single parallel-jaw gripper can, with no new target data and no retraining, act on end-effectors not seen during training.
Unique to our method compared to our baselines are the wrist masking strategy and augmentations.
Therefore, our experiments focus on differences in the masking protocol, or lack thereof.
Specifically, our experiments aim to address the following questions:

\begin{itemize}[nosep, leftmargin=0em, label={}]
  \item \textbf{(Q1)} Does cloaking the wrist view degrade the source embodiment performance?
  \item \textbf{(Q2)} Does \vlaname{} transfer zero-shot to an unseen gripper, arm, and a multi-finger hand?
  \item \textbf{(Q3)} Is masking the wrist view the critical enabler of cross-embodiment transfer?
  \item \textbf{(Q4)} Is our mask augmentation strategy (\S~\ref{sec:method:mask}) necessary, and how does it compare to alternatives?
\end{itemize}

\subsection{Experimental Setup}
\label{sec:exp:setup}
\paragraph{Data and training.}
We train on DROID~\citep{khazatsky2024droid, pertsch2024droidlang}, a large-scale in-the-wild teleoperation dataset of $\sim\!75$K episodes ($\sim\!350$ hours) collected on a Franka Panda arm with a Robotiq 2F-85 parallel-jaw gripper as the single source embodiment.
We train \vlaname{} by finetuning \baselinePi with \methodname{} on DROID, following an open-source post-training recipe~\cite{openpi}. We train for $100$k gradient steps, or approximately 1 epoch, requiring 48 hours on 8 H100 GPUs on Stanford's Marlowe cluster~\cite{marlowe2025}.

\paragraph{Tasks and protocol.}
We evaluate each baseline on four common tasks in DROID, ``Pick and Place'', ``Remove'', ``Move'', and ``Fold/Unfold'', each requiring accurate tip placement and a firm grasp, the regime our tip-pose retargeting addresses.
We deploy on four embodiments. Three are mounted on the same Franka arm: the Robotiq 2F-85 from the DROID dataset, an unseen UMI gripper that differs in color, shape, and opening size, and the Sharpa five-fingered humanoid hand. We also deploy on a YAM arm and gripper.
For the Sharpa, we designed a camera mount with an offset that places the fingers at roughly the same image location as the original gripper.
For each task, we generate 12 randomized instances of objects, placements, and prompts, for a total of 48 trials. Instances are generated once and reused identically across all baselines. Each trial is run for a maximum of 60\,s or terminated early when it risks damaging the end-effector.
Our primary metric is task progression rate, where increased score is obtained for completing subtasks that make progress toward the goal.
Full definitions, success criteria, and object sets are detailed in Section~\ref{subsec:supp_task_definitions}.

\paragraph{Baselines.}
We compare our policy, \vlaname{}, against four baselines. \textbf{\baselinePiDroid} is $\pi_{0.5}$ finetuned on DROID, which amounts to our method without any masking but with tip-pose retargeting (TP retargeting). \textbf{\baselineNoAug} is our method without the mask augmentations described in Section~\ref{sec:method:mask}. \textbf{\baselineOverlap}, inspired by Shadow~\citep{lepert2024shadow}, overlays source and target masks during training rather than the augmentation strategy in Section~\ref{sec:method:mask}. We also compare to \textbf{\baselineLAP}~\citep{zha2026laplanguageactionpretrainingenables}, a recent zero-shot cross-embodiment policy that aligns embodiments at the language layer. We additionally ablate the wrist camera itself with \textbf{\baselinePiDroidNoWrist}, \baselinePiDroid trained on the external view only, evaluated on the source gripper and the Sharpa hand (Table~\ref{tab:no_wrist_ablation}). See Section~\ref{subsec:supp_baselines} for additional baseline details and a discussion of their fairness.

\subsection{Results}

\begin{figure}[t]
    \centering
    \includegraphics[width=\columnwidth]{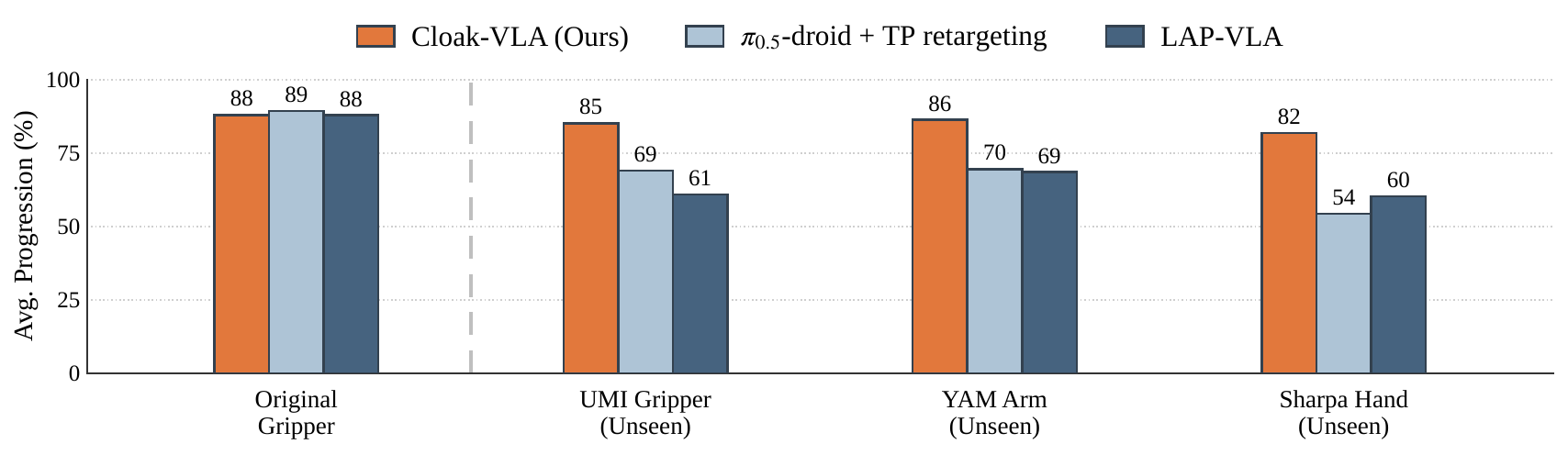}
    
    \vspace{-5pt}
    \caption{Task-averaged progression rate. The original gripper does not use TP retargeting because there is no embodiment transfer. All methods tie on the source gripper. On the unseen YAM arm, UMI gripper, and Sharpa hand, only \methodname{} holds up while every baseline falls behind, most notably on the hand. }
    \vspace{-5pt}
    \label{fig:main_result_avg}
\end{figure}

\begin{table}[t]
\centering
\small
\setlength{\tabcolsep}{6pt}
\renewcommand{\arraystretch}{1.2}
\resizebox{\textwidth}{!}{%
\begin{tabular}{l l ccccc}
\toprule
 & & \multicolumn{4}{c}{Progression rate (\%) $\uparrow$} & \\
\cmidrule(lr){3-6}
Embodiment & Method & Pick and Place & Remove & Move & Fold or Unfold & Task Average \\
\midrule
\multirow{6}{2cm}{\textbf{Original Gripper}}
    & \baselinePiDroid\ver{v0} & \textbf{\val{100.0}{0.0}} & \textbf{\val{100.0}{0.0}} & \val{70.8}{10.1} & \val{86.2}{4.9} & \textbf{\val{89.3}{3.2}} \\
    & \baselineLAP & \val{95.8}{2.8} & \val{83.2}{8.7} & \textbf{\val{83.3}{8.9}} & \val{\underline{89.0}}{4.7} & \val{87.9}{3.4} \\
    & \baselineNoAug\ver{v3.0} & \val{\underline{97.9}}{2.1} & \val{\underline{94.4}}{5.6} & \val{\underline{72.9}}{11.7} & \val{77.9}{6.3} & \val{85.8}{3.8} \\
    & \baselineOverlap\ver{v3.2} & \val{83.3}{7.7} & \val{\underline{94.4}}{5.6} & \val{70.8}{9.1} & \val{77.9}{8.5} & \val{81.6}{4.0} \\
    & \textbf{\vlaname{} (ours)}\ver{v3.1} & \val{\underline{97.9}}{2.1} & \val{91.7}{6.0} & \val{70.8}{9.6} & \textbf{\val{91.8}{4.3}} & \val{\underline{88.0}}{3.3} \\
 
\midrule

\multirow{6}{2cm}{\textbf{UMI Gripper}\\(Unseen)}
    & \baselinePiDroid\ver{v0} + TP retargeting & \val{\underline{95.8}}{2.8} & \val{69.3}{9.6} & \val{43.8}{9.8} & \val{67.0}{0.0} & \val{69.0}{4.3} \\
    & \baselineLAP  & \val{70.8}{8.6} & \val{52.7}{11.2} & \val{56.2}{8.8} & \val{64.0}{8.7} & \val{60.9}{4.7} \\
    & \baselineNoAug\ver{v3.0} & \val{89.6}{4.8} & \val{\underline{88.8}}{7.5} & \val{\underline{70.8}}{10.1} & \textbf{\val{83.5}{5.0}} & \val{\underline{83.2}}{3.7} \\
    & \baselineOverlap\ver{v3.2} & \val{93.8}{6.2} & \val{86.1}{7.7} & \val{\underline{70.8}}{9.6} & \val{\underline{77.9}}{6.3} & \val{82.1}{3.9} \\
    & \textbf{\vlaname{} (ours)}\ver{v3.1} & \textbf{\val{97.9}{2.1}} & \textbf{\val{91.7}{6.0}} & \textbf{\val{72.9}{8.4}} & \val{77.8}{9.5} & \textbf{\val{85.1}{3.7}} \\

\midrule

\multirow{6}{2cm}{\textbf{YAM Arm \\ and Gripper}\\(Unseen)}
    & \baselinePiDroid\ver{v0} + TP retargeting & \val{68.8}{10.3} & \val{\underline{80.5}}{8.7} & \val{47.9}{10.0} & \val{80.7}{6.4} & \val{69.5}{4.8} \\
    & \baselineLAP  & \val{77.1}{6.5} & \val{72.2}{8.1} & \val{60.4}{8.9} & \val{64.1}{5.0} & \val{68.5}{3.7} \\
    & \baselineNoAug\ver{v3.0} & \val{\underline{91.7}}{6.4} & \val{69.4}{10.5} & \textbf{\val{83.3}{7.7}} & \val{66.8}{9.2} & \val{\underline{77.8}}{4.4} \\
    & \baselineOverlap\ver{v3.2} & \val{58.3}{11.2} & \val{\underline{80.5}}{8.7} & \val{62.5}{7.2} & \textbf{\val{86.2}{6.4}} & \val{71.9}{4.5} \\
    & \textbf{\vlaname{} (ours)}\ver{v3.1} & \textbf{\val{93.8}{3.3}} & \textbf{\val{97.2}{2.8}} & \val{\underline{70.8}}{11.9} & \val{\underline{83.4}}{6.5} & \textbf{\val{86.3}{3.7}} \\

\midrule

\multirow{6}{2cm}{\textbf{Sharpa Hand}\\(Unseen)}
    & \baselinePiDroid\ver{v0} + TP retargeting & \val{47.9}{8.9} & \val{72.4}{5.6} & \val{33.3}{10.4} & \val{64.1}{7.6} & \val{54.4}{4.6} \\
    & \baselineLAP  & \val{\underline{81.2}}{9.8} & \val{47.1}{8.7} & \val{54.2}{9.6} & \val{58.3}{7.3} & \val{60.2}{4.7} \\
    & \baselineNoAug\ver{v3.0} & \val{64.6}{10.4} & \val{\underline{80.7}}{6.4} & \val{45.8}{11.9} & \val{69.5}{7.7} & \val{65.1}{4.9} \\
    & \baselineOverlap\ver{v3.2} & \val{75.0}{10.2} & \val{74.9}{9.3} & \textbf{\val{68.8}{10.7}} & \val{\underline{69.6}}{8.7} & \val{\underline{72.1}}{4.7} \\
    & \textbf{\vlaname{} (ours)}\ver{3.1}   & \textbf{\val{93.8}{3.3}} & \textbf{\val{86.1}{7.7}} & \val{\underline{66.7}}{10.4} & \textbf{\val{80.7}{6.4}} & \textbf{\val{81.8}{3.9}} \\
    
\bottomrule
\end{tabular}%
}
\caption{Zero-shot transfer across embodiments. Each cell reports task progression rate (mean $\pm$ SEM, \%). \vlaname{} is trained only on the Original Gripper and deployed zero-shot to UMI, Sharpa, and YAM. For each embodiment, \textbf{bold} marks the best score in the column and \underline{underline} marks the second-best.
}
\label{tab:results}
\end{table}

Quantitative results are reported in Table~\ref{tab:results} and Figure~\ref{fig:main_result_avg}. For qualitative cross-embodiment rollouts, including common fail cases, we refer the reader to our project site and Section~\ref{sec:supp_results}.

\paragraph{Source embodiment (Q1).}
On the source Robotiq gripper, where there is no transfer gap, we compare \vlaname{} against the standard \baselinePiDroid. The only difference between the two is the use of \methodname{} masking in the wrist camera.
\vlaname{} leads on Fold/Unfold and ties on Move, while \baselinePi is slightly ahead on Pick and Place and Remove. On average, the two methods perform equally, within error margin.
\methodname{} masks patches from the wrist view that carry little information about the surrounding scene. Cloaking them, therefore, can be done at no cost on the source embodiment.
However, if more patches are masked than necessary, as in \baselineOverlap, scene information is lost and performance slightly degrades.

\paragraph{Zero-shot transfer (Q2, Q3).}
Zero-shot transfer to an unseen embodiment incurs an inevitable performance cost; the methods differ in its magnitude (Figure~\ref{fig:main_result_avg}). Across all three unseen embodiments, \vlaname{} remains close to the source embodiment progression rate of $88.0$, to $85.1$ on the UMI gripper, $86.3$ on the YAM arm, and $81.8$ on the Sharpa hand. Both \baselinePiDroid and \baselineLAP degrade significantly.
The contrast is sharpest on the Sharpa hand; \vlaname{} drops slightly from its source embodiment performance of $88.0$ to $81.8$, while \baselinePiDroid drops significantly from $89.3$ to $54.4$ and \baselineLAP from $87.9$ to $60.2$.
\vlaname{} performs the best on every unseen embodiment, demonstrating that \methodname{} enables zero-shot transfer to embodiments unseen during training (Q2).
Moreover, addressing the vision gap is the critical enabler of the above zero-shot transfer.
\baselinePiDroid differs from \vlaname{} only in the wrist masking, and \baselineLAP aligns embodiments at the language-action interface while leaving the wrist view untouched. Both degrade substantially on the unseen embodiments, while \vlaname{} holds up, indicating that masking the wrist view, rather than aligning the action interface, carries the policy across embodiments (Q3).

\begin{table}[t]
\centering
\small
\setlength{\tabcolsep}{6pt}
\renewcommand{\arraystretch}{1.2}
\resizebox{\textwidth}{!}{%
\begin{tabular}{l l ccccc}
\toprule
 & & \multicolumn{4}{c}{Progression rate (\%) $\uparrow$} & \\
\cmidrule(lr){3-6}
Embodiment & Method & Pick and Place & Remove & Move & Fold or Unfold & Task Average \\
\midrule
\multirow{3}{2cm}{\textbf{Original Gripper}}
    & \baselinePiDroid & \textbf{\val{100.0}{0.0}} & \textbf{\val{100.0}{0.0}} & \textbf{\val{70.8}{10.1}} & \val{86.2}{4.9} & \textbf{\val{89.3}{3.2}} \\
    & \baselinePiDroidNoWrist & \val{37.5}{10.4} & \val{33.1}{7.1} & \val{22.9}{4.8} & \val{52.7}{9.6} & \val{36.5}{4.3} \\
    & \textbf{\vlaname{} (ours)} & \val{97.9}{2.1} & \val{91.7}{6.0} & \textbf{\val{70.8}{9.6}} & \textbf{\val{91.8}{4.3}} & \val{88.0}{3.3} \\
\midrule
\multirow{3}{2cm}{\textbf{Sharpa Hand}\\(Unseen)}
    & \baselinePiDroid + TP retargeting & \val{47.9}{8.9} & \val{72.4}{5.6} & \val{33.3}{10.4} & \val{64.1}{7.6} & \val{54.4}{4.6} \\
    & \baselinePiDroidNoWrist + TP retargeting & \val{39.6}{7.8} & \val{38.8}{9.9} & \val{22.9}{7.8} & \val{50.0}{7.7} & \val{37.8}{4.3} \\
    & \textbf{\vlaname{} (ours)} & \textbf{\val{93.8}{3.3}} & \textbf{\val{86.1}{7.7}} & \textbf{\val{66.7}{10.4}} & \textbf{\val{80.7}{6.4}} & \textbf{\val{81.8}{3.9}} \\
\bottomrule
\end{tabular}%
}
\caption[No-wrist ablation.]{No-wrist ablation. Task progression rate (mean $\pm$ SEM, \%) on the source gripper and the unseen Sharpa hand. \baselinePiDroidNoWrist is \baselinePiDroid trained on the external camera only. Removing the wrist view collapses performance on both embodiments, whereas \methodname{} keeps the wrist view and cloaks only the end-effector.}
\label{tab:no_wrist_ablation}
\end{table}

\paragraph{No-wrist ablation (Q3).}
A simpler way to hide the end-effector from the policy is to omit the wrist camera entirely. We test this with \baselinePiDroidNoWrist, which is \baselinePiDroid trained on the external view only (Table~\ref{tab:no_wrist_ablation}). Performance collapses on both the source gripper (from $89.3$ to $36.5$) and the Sharpa hand (from $54.4$ to $37.8$), confirming that the external camera carries little task signal on its own. This motivates our decision to mask only the wrist view and leave the external camera untouched.

\paragraph{Mask augmentation ablation (Q4).}
We ask whether our mask augmentation strategy (\S~\ref{sec:method:mask}) is necessary and how it compares to alternatives.
We ablate our strategy against \baselineNoAug, which trains on the unaugmented source mask, and \baselineOverlap, inspired by Shadow~\cite{lepert2024shadow}, which overlays the target mask during training and therefore requires the target embodiment to be known a priori.
Across all three unseen embodiments, \vlaname{} outperforms both ablations (Table~\ref{tab:results}). The margin is widest on the Sharpa hand, where \vlaname{} reaches $81.8$ versus $65.1$ for \baselineNoAug and $72.1$ for \baselineOverlap.
This indicates that mask augmentation is necessary, and that augmenting the source silhouette yields better generalization than overlaying a known target mask.
We attribute this generalization to our strategy exposing the model to a distribution of end-effector silhouettes, whereas training on a fixed silhouette is brittle.

\section{Conclusion}
\label{sec:conclusion}

\methodname{} shows that a single-embodiment policy can be decoupled from its own body by cloaking the embodiment in the wrist view, resulting in visual and spatial reasoning that is largely embodiment-agnostic.
Beyond zero-shot deployment, the resulting body-agnostic model is also a strong initialization for finetuning or continual learning on a new embodiment.
Data gathered on one end-effector is therefore not thrown out when the hardware changes; it can seed policies for future embodiments.

\textbf{Limitations and future work.}
Our approach covers manipulation expressible via two fingertips and not skills that demand richer contact or in-hand reorientation. The behavior we show is inherited from the source model and the gripper-style tabletop data in DROID, a limitation that richer data would mitigate. Accordingly, the five-fingered hand is used as a two-tip gripper here, and we do not claim dexterous skills.
The tip-pose retargeting also introduces limitations. To avoid instability in the IK null space solutions, we introduce regularizing objectives that suppress tilting behaviors that would better suit certain tasks. This is a tradeoff that per-embodiment tuning could relax, but puts a limitation on cross-embodiment generalization.
The mask also acts at the encoder's patch resolution and can occlude parts of an object near the end-effector; a finer patch resolution would reduce occlusion.
Finally, transfer is not lossless; closing the residual gap with a few demonstrations from our embodiment-agnostic checkpoint is a natural next step.

\acknowledgments{This work was supported by the National Science Foundation under Grant No.~2153854, the Stanford Institute for Human-Centered AI (HAI), and the Stanford Wu Tsai Human Performance Alliance. We thank Albert Wu and Sarthak Kamat for their technical guidance and insightful discussions, and Haochen Shi for designing the camera mounts.}

\bibliography{example}  %

\ifmodefull
\clearpage
\appendix

\fi

\fi %

\end{document}